\documentclass[submission,copyright,creativecommons]{eptcs}
\providecommand{\event}{ICLP 2021}
\usepackage{underscore}           

\def\easp{{\tt exp(ASP$^c$)}}
\def\oldeasp{{\tt exp(ASP)}}
\usepackage{amsmath,amsfonts,amssymb}
\usepackage{graphicx}
\usepackage{graphicx,picinpar,wrapfig, blindtext}
\usepackage{ifthen}
\usepackage[table]{xcolor}
\def\naf{ \: not \: } 
\definecolor{Gray}{rgb}{0.47,0.53,0.6}
\definecolor{LightCyan}{rgb}{0.88,1,1}
\definecolor{yellow}{rgb}{1,1,0.88}
\definecolor{azure}{rgb}{0.94,1,1}
\definecolor{lightgrey}{rgb}{0.83,0.83,0.83}
\newcommand{\uffa}{\mbox{$\: {\tt : \!\! - }\:$}} 
\usepackage{multicol}
\usepackage{xcolor}
\usepackage[linesnumbered,ruled,vlined]{algorithm2e}

\SetCommentSty{mycommfont}
\usepackage{enumitem}

\SetKwInput{KwInput}{Input}                
\SetKwInput{KwOutput}{Output}              
\SetKw{Break}{break}
\SetKw{Go}{goto line}
\usepackage{lipsum}

\usepackage{listings}
\lstdefinelanguage{clingo}{
  keywordstyle=[1]\usefont{OT1}{cmtt}{m}{n},%
  keywordstyle=[2]\textbf,%
  keywordstyle=[3]\usefont{OT1}{cmtt}{m}{n},
  alsoletter={\#,\&},%
  keywords=[1]{not,from,import,exists,if,else,return,while,
		break,and,or,for,in,del,and,class,subClass,concern,aspect,subCo,prop,rdf,cpsf,addBy,suppBy,neg,d,relation,holds,h,obs,action,fluent,occurs,req,group,leadTo,active,deg,comp,order,hSubCo,llh_sat_sub,llh_sat,llh_sat_sub_aux,step,deg_pos,nAllPosCon,nActPosCon,possImpactsPos,scoreLoS,wBus,wHum,wTru,wFun,wTim,wBou,wLif,wCom,wDat,last,addFun,func,formula,sat_formula,possImpactsNeg,conflict,member,sa_action,exec,sc_concern},%
  keywords=[2]{\#const,\#show,\#minimize,\#maximize,\#base,\#theory,\#count,\#external,\#program,\#script,\#end,\#heuristic,\#edge,\#project,\#show},%
  keywords=[3]{&,&dom,&sum,&diff,&show,&minimize},%
  morecomment=[l]{\#\ },%
  morecomment=[l]{\%\ },%
  commentstyle={\color{darkgray}}%
}
\newtheorem{example}{Example}  
\def\naf{ \: not \: } 

\title{exp(ASP$^c$): Explaining ASP Programs with Choice Atoms and Constraint Rules\thanks{The second would like to acknowledge the partial support of the NSF 1812628 grant.}}
\author{Ly Ly Trieu \qquad\qquad Tran Cao Son
\institute{New Mexico State University\\
New Mexico, USA}
\email{lytrieu@nmsu.edu \qquad\qquad tson@cs.nmsu.edu}
\and
Marcello Balduccini 
\institute{Saint Joseph's University\\
Pennsylvania, USA}
\email{\quad mbalducc@sju.edu}
}
\def\titlerunning{exp(ASP$^c$): Explaining ASP Programs with Choice Atoms and Constraint Rules}
\def\authorrunning{L.L. Trieu, T.C. Son \& M. Balduccini}
\begin{document}
\maketitle

\begin{abstract}
We present an enhancement of \oldeasp{}, a system that generates explanation graphs for a literal $\ell$---an atom $a$ or its default negation $\sim{a}$---given an answer set $A$ of a normal logic program $P$, which explain why $\ell$ is true (or false) given $A$ and $P$. The new system, \easp{}, differs from \oldeasp{} in that it supports choice rules and utilizes constraint rules to provide explanation graphs that include information about choices and constraints.  
\end{abstract}

\section{Introduction}

\emph{Answer Set Programming (ASP)} \cite{MarekT99,Niemela99} is a popular paradigm for decision making and problem solving in Knowledge Representation and Reasoning. 
It has been successfully applied in a variety of applications such as robotics, planning, diagnosis, etc. ASP is an attractive programming paradigm as it is a declarative language, where programmers focus on the representation of a specific problem as a set of rules in a logical format, and then leave computational solutions of that problem to an answer set solver. However, this mechanism typically gives little insight into \emph{why} something is a solution and \emph{why} some proposed set of literals is not a solution. This type of reasoning falls within the scope of \emph{explainable Artificial Intelligence} and is useful to enhance the understanding of the resulting solutions as well as for debugging programs. So far, only a limited number of approaches have been proposed
\cite{Cabalar2020,PontelliSE09,schulz2016justifying}. To the best of our knowledge, no system deals directly with ASP programs with choice atoms.  

\begin{figwindow}[0,r,%
{
\includegraphics[width=0.35\textwidth]{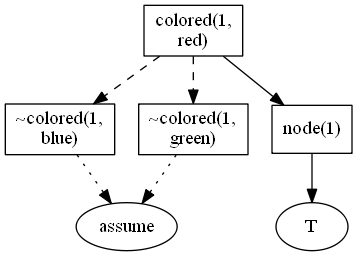} 
},%
{\scriptsize Explanation of  $colored(1,red)$ \label{fig:graphcoloring}}] 
In this paper, we present an improvement over our previous system, \emph{exp(ASP)} \cite{lyly2021}, called \easp{}. Given an ASP program $P$, an answer set $A$, and an atom $a$, \easp{} is aimed at answering the question ``why is $a$ true/false in $A$?'' by producing \emph{explanation graphs} for atom $a$. The current system, \texttt{exp(ASP)}, does not consider programs with choice atoms and other constructs that extend the modeling capabilities of ASP. For instance, Fig.~\ref{fig:graphcoloring} shows an explanation graph for the atom $colored(1,red)$ given the typical encoding of the graph coloring problem that does not use choice rules. This explanation graph does provide the reason for the color assigned to node 1 by indicating that the node is red because it is not blue and not green. It is not obvious that this information represents the requirement that each node is colored with exactly one color. The improvement described here extends our approach with the ability to handle ASP programs containing choice rules and include constraint information in the explanation graphs. 
\end{figwindow} 

\section{Background: The \texttt{exp(ASP)} System}
\label{subsec:background:exp(ASP)}

\oldeasp{} deals with normal logic programs which are collection of rules of the form 
$head(r) \leftarrow body(r)$  
where $head(r)$ is an atom and 
$body(r) = r^+, \naf r^-$ with $r^+$ and $r^-$ collections of atoms  
in a propositional language and $\naf r^-$ denotes the set 
$\{\naf x \mid x \in r^-\}$ 
and $\mathit{not}$ is the default negation.  

\oldeasp{} generates explanation graphs under the answer set semantics \cite{GelfondL88}.
It implemented the algorithms proposed in \cite{PontelliSE09} to generate explanation graphs of a literal  $\ell$ ($a$ or $\sim{a}$ for some atom $a$ in the Herbrand base $H$ of $P$), given an answer set $A$ of a program $P$. Specifically, the system produces labeled directed graphs, called \emph{explanation graphs}, for $\ell$, whose nodes belong to $H \cup \{\sim{x} \mid x \in H\} \cup \{\top,\bot,\mathtt{assume}\}$ and whose links are labeled with $+$, $-$ or $\circ$ (in Fig.~\ref{fig:graphcoloring}, solid/dash/dot edges represent $+$/$-$/$\circ$ edges). 
Intuitively, for each node $x$, $x\not\in\{\top,\bot,\mathtt{assume}\}$ in an explanation graph $(E,G)$, the set of neighbors of $x$ represents a support for $x$ being 
true given $A$ (see below). 

The main components of \oldeasp{} are:
\begin{enumerate}
    \item \textbf{Preprocessing:} This component produces an $\mathit{aspif}$ representation \cite{kaminski2017tutorial} of $P$ that will be used in the reconstruction of ground rules of $P$. It also computes supported sets for atoms (or its negations) 
    in the Herbrand base of $P$ and stored in an associative array $E$.

    \item \textbf{Computing minimal assumption set:} This calculates a minimal assumption set $U$ given the answer set $A$ and $P$ according to the definition in \cite{PontelliSE09}.
    
    \item \textbf{Computing explanation graphs:} This component uses the supported sets in $E$ and constructs e-graphs for atoms in $H$ (or their negations) under the assumption that each element $u \in U$ is assumed to be false.  
    
\end{enumerate}

We note that \oldeasp{} does not deal with choice atoms  \cite{simons2002extending}.
The goal of this paper is to extend \oldeasp{} to deal with choice atoms and utilize constraint information.

\section{Explanation Graphs in Programs with Choice Atoms}

\oldeasp{} employs the notion of a \emph{supported set} of a literal in a program in its construction. Given a program $P$, an answer set $A$ of $P$, and an atom $c$, if $c\in A$ and $r$ is a rule such that (\emph{i}) $head(r) = c$, 
 (\emph{ii}) $r^+ \subseteq A$, and (\emph{iii}) 
 $r^- \cap A = \emptyset$, $support(c,r) = r^+ \cup \{\sim\!n \mid n \in r^-\}$; 
 and refer to this set as a \emph{supported set} of $c$ for rule $r$. 
 If $c \notin A$, for every rule $r$ such that $head(r)=c$, then  
$support(\sim\!c,r) \in \{\{p\} \mid p \in A \cap r^- \} \cup \{\{\sim\!n \} \mid n \in r^+\setminus A\}$.  

To account for choice atoms\footnote{We use choice atoms synonymous with weight constraints.} in $P$, the notion of supported set needs to be extended. For simplicity of the presentation, we assume that any choice atom $x$ is of the form $l~ \{p_1 :q_1, \ldots, p_n:q_n\} ~u$  where\footnote{As we employ the \emph{aspif} representation, this is a reasonable assumption.}  
$p_i$'s and $q_i$'s  are atoms. Let $x_l$ and $x_u$ denote $l$ and $u$, respectively. Furthermore, we write $c \in x$ to refer to 
an element in $\{p_1,\ldots,p_n\}$. For $c \in x$,  
$q_i \cong c$ indicates that $c : q_i$ belongs to $\{p_1 :q_1, \ldots, p_n:q_n\}$.  
 
In the presence of choice atoms, an atom $c$ can be true because 
$c$ belongs to a choice atom that is a head of a rule $r$ and 
$body(r)$ is true in $A$. In that case, we say that $c$ is chosen to be true 
and extend $support(c,r)$ with a special atom ${+}choice$ to indicate that 
$c$ is chosen to be true. Likewise, $c$ can be false even if it belongs to a choice atom that is a head of a rule $r$ and 
$body(r)$ is true in $A$. In that case, we say that $c$ is chosen to be false 
and extend $support(\sim{c},r)$ with a special atom ${-}choice$ to indicate that 
$c$ is chosen to be false.
Also, $q \cong c$ will belong to the support set of $support(c,r)$ and $support(\sim{c},r)$.

The above extension only considers the case $c$ belongs to the head of a rule. $support(c,r)$ also needs to be extended with atoms corresponding to choice atoms in the body of $r$.  Assume that $x$ is a choice atom in $r^+$. By definition, if $body(r)$ is true in $A$ then  $x_l \le |S| \le x_u$ where $S = \{(c,q) \mid c \in x ,q \cong c, A \models c \land q  \}$. For this reason, we extend $support(c,r)$ with $x$. 
Because $x$ is not a standard atom, we indicate the support of $x$ given $A$ by 
defining $support(x,r) = \{S\}$. Furthermore, for each $s \in S$, 
$support(s,r) = \{*True\}$. When $S=\emptyset$, we write $support(x,r) = \{*Empty\}$.
Similar elements will be added to $support(c,r)$ or $support(\sim{c},r)$ in other cases (e.g.,
the choice atom belongs to $r^-$) or has different form (e.g., when $l=0$ or $u=\infty$). We omit the precise definitions of the elements that need to be added to $support(c,r)$ for brevity.

The introduction of different elements in supported sets of literals in a program necessitates the extension of the notion of explanation graph. Due to the space limitation, we introduce its key components and provide the intuition behind each component. The precise definition of an explanation graph is rather involved and is included in the appendix for review. First, we introduce additional types of nodes. Besides $\mathtt{+choice}$, $\mathtt{-choice}$, $\mathtt{*True}$, and $\mathtt{*Empty}$, we consider the following types of nodes:  

\begin{list}{$\bullet$}{\itemsep=0pt \topsep=1pt  \parsep=0pt}
    \item \emph{Tuples} are of the form $(x_1,\ldots, x_m, \naf y_1,\ldots, \naf y_n)$
    to represent elements belonging to choice atoms 
    (e.g., $(colored(1,blue),color(blue))$ representing 
    an element in $1 \{(colored(N,C) : color(\\C)\} 1$). 
    $\mathcal{T}$ denotes all tuple nodes in program $P$. 
    
    \item \emph{Choices} are of the form $l \le T \le u$ or $\sim{(l \le T \le u)}$ 
    where $T \subseteq \mathcal{T}$. Intuitively, when  $l \le T \le u$ (resp. $\sim{(l \le T \le u)}$) occurs in an explanation graph, it indicates that  
    $l \le T \le u$ is satisfied (resp. not satisfied) in the given answer set $A$.  
    $\mathcal{O}$ denotes all choices.  
    
    \item \emph{Constraints} are of the form $triggered\_constraint(x)$ 
    or $triggered\_constraint(\sim{x})$. The former (resp. latter) indicates that $x$ is (resp. is not) true in $A$ and satisfies all the constraints $r$ such that $x \in r^+$  (resp. $x \in r^-$). 
    The set of all constraints is denoted with $\mathcal{C}$.
\end{list}
As defined the nodes of the graph, we introduce the new types of links in explanation 
graphs as follows:
\begin{list}{$-$}{\itemsep=0pt \topsep=1pt  \parsep=0pt}
    \item  $\bullet$ is used to connect literals $c$ and $\sim\!c$ to $\mathtt{+choice}$ and $\mathtt{-choice}$, respectively, where $c \in x$ and $x$ is a choice atom in the head of a rule. 
    \item  $\diamond$ is used to connect literals $c$ and $\sim\!c$ to $triggered\_constraint(c)$ and $triggered\_constraint(\sim\!c)$, respectively.
    \item  $\oplus$ is used to connect a tuple $t \in \mathcal{T}$ to $\mathtt{*True}$.
    \item  $\oslash$ is used to connect a choice $n \in \mathcal{O}$ to $\mathtt{*Empty}$.
\end{list}

\section{The \texorpdfstring{\easp{}} System}
\label{sec:system}
In this section, we will focus on describing how the three main tasks in Sec.\ \ref{subsec:background:exp(ASP)} are implemented. {\tt exp(ASP\\$^c$)} uses a data structure, associative array, whose keys can be choices, tuples, constraints, or literals.
For an associate array $D$, we use $D.keys()$ to denote the set of keys in $D$ and 
$k \mapsto D[k]$ to denote that $k$ is associated to $D[k]$. 
To illustrate the different concepts, we will use the program $P_1$ that contains a choice atom and a constraint rule as follows:
\[\begin{array}{clclcclclcclcl} 
(r_1) & \texttt{a} & \uffa & \:\:not\: \texttt{b}, \:\:not\:\texttt{c}. & \hspace{1cm} & (r_3) & \texttt{c} & \uffa & \:\:not\: \texttt{a}.  & \hspace{1cm} &   (r_5) & \texttt{1 \{m(X) : n(X)\} 1} & \uffa & \texttt{c}. \\
(r_2) & \texttt{b} & \uffa & \texttt{c}, \texttt{a}. & & (r_4)  &  & \uffa & \texttt{b}, \texttt{m(1)}.& \hspace{1cm} & (r_6)  & \texttt{n(1..2)}.\\
  \end{array}
\]

\subsection{Preprocessing}
\label{subsec:preprocessing}

Similar to \oldeasp{}, a program is preprocessed to maintain facts and as many ground rules as possible by using the {\small \tt --text} and {\small \tt --keep-facts}  options and replacing facts with the external statements. The $\mathit{aspif}$ representation of the program is then obtained and processed, together with the given answer set, for generating explanation graphs. The $\mathit{aspif}$ statements of $P_1$ is given in Listing \ref{lst:ex1}. Let us briefly discuss the $\mathit{aspif}$ representation before continuing with the description of other components.

\begin{wrapfigure}[24]{l}{.42\textwidth}
\begin{lstlisting}[language=clingo,caption=$\mathit{aspif}$ Representation of $P_1$, label=lst:ex1, mathescape=true,xleftmargin=.1\textwidth, breaklines=true, numberstyle=\tiny]
asp 1 0 0
5 1 2
5 2 2
1 0 1 3 0 1 -4
1 0 1 4 0 2 -5 -3
1 0 1 5 0 2 4 3
1 0 1 6 0 1 3
1 0 0 0 2 7 5
1 1 1 7 0 2 6 1
1 1 1 8 0 2 6 2
1 0 1 9 0 2 1 7
1 0 1 10 0 2 2 8
1 0 1 11 1 1 2 9 1 10 1
1 0 1 12 1 2 2 9 1 10 1
1 0 1 13 0 2 11 -12
1 0 0 0 2 6 -13
4 4 n(1) 1 1
4 4 n(2) 1 2
4 1 b 1 5
4 1 c 1 3
4 1 a 1 4
4 4 m(1) 1 7
4 4 m(2) 1 8
0
\end{lstlisting}
\end{wrapfigure}

Each line encodes a statement in \emph{aspif}. Lines starting with 4, 5, and 1 are  
output, external, and rule statements, respectively.   
Atoms are associated with integers and encoded in output statements (e.g., Line 17: $1$ 
is the identifier of $n(1)$). External statements help us to recognize the facts in $P$, e.g. atom $n(1)$ ($ID = 1$) is a fact (Line 2). A rule statement $r$ is of the form: $1 ~ H ~B$, where $H$ and $B$ are the encoding of the head and body of $r$, respectively.
Because of page limitation, we focus on describing the rule statement whose head is a choice atom or whose body is a weight body.
If the head is a choice, its encoding $H$ has the form: 
    $1 ~ n ~ i_{c_1} ~ \ldots ~i_{c_n}$, where $n$ is the number of head atoms and $i_c$ is an integer identifying the atom $c$.  
E.g. $m(1)$ ($ID = 7$) and $m(2)$ ($ID = 8$) are the head choices in Lines 9 and 10, respectively, which represents rule $r_5$.
If the body of a rule is a weight body, its encoding $B$ has the form:
    $1 ~ l ~ n ~i_{a_1} ~ w_{a_1} ~ \ldots ~ i_{a_n} ~ w_{a_n}$, where $l > 0, l \in \mathbb{N}$ is the lower bound, $n > 0$ is the number of literal ${a_i}$'s with $ID = i_{a_i}$ and weight $w_{a_i}$. E.g. Lines 13-14 contain weight body. 
Given an ID $i$ that does not occur in any output statement 
\cite{kaminski2017tutorial,lyly2021}, we use $l(i)$ to denote the corresponding literal.
Constraint $r_4$ is shown in Line 8.
It is interesting to observe that there is one additional constraint in Line 16.
By tracking integer identifiers, one can notice that Line 16 states that it cannot be the case that $c$ is true (via Line 7) and $l(13)$ cannot be proven to be true. Lines 13-15 ensure that $l(13)$ is true if $1 \{ l(9); l(10)\}$ is true and $2 \{l(9); l(10)\}$ cannot be proven to be true. Note that $l(9)$ and $l(10)$ are the same weight, so we ignore their weight. 
Line 11 states that $l(9)$ is true if $m(1)$ and $n(1)$ are true. 
Line 12 states that $l(10)$ is true if $m(2)$ and $n(2)$ are true. 
Thus, the new constraint is generated from the semantics of choice rule $r_5$, which is added to aspif representation.

Given the $\mathit{aspif}$ representation $P'$ of a program $P$, 
an associate array $D_P$ is created where   
$D_P = \{(t,h) \mapsto B \mid t \in \{0,1\}, h \in H, B=\{body(r) \mid r \in P', head(r)=h\}\}$. Here, for an element $(t,h) \mapsto B$ in $D_p$, $t$ is the type of head $h$, either disjunction ($t = 0$) or choice ($t=1$). 
Furthermore, for an answer set $A$ of $P$, 
$E_{r(P)} = \{k \mapsto V \mid k \in \{a \mid a \in A\} \cup \{\sim\!a \mid a \notin A\},  V = \{support(k,r) \mid r \in P\}\}$ \cite{lyly2021}.

Algorithm~\ref{alg:constraintpreprocessing} shows how constraints are processed given 
the program $P$ and its answer set $A$. The outcome of this algorithm is an associated array $E_c$. First, $V_c$---the set of constraints (the bodies of constraints)---is computed.  
Afterwards, for each body $B$ of a constraint $r$ in $V_c$, $violation$ 
and $support$ are computed. 
Each element in $violation$ requires some trigger constraint to falsify the body $B$, which are those in $support$. Each $choice\_support$ encodes a support for a choice atom. $triggered\_constraint(v)$, where $ v \in violation$, is assigned to support the explanation of $v$ (Line~\ref{pre:triggeredassign}), and $support$ is used for the explanation of $triggered\_constraint(v)$ to justify the satisfaction of constraints containing $v$ (Line~\ref{pre:triggeredsupport}).
For $\{p_1 :q_1, \ldots, p_n:q_n\}$ in a choice atom $x$, we write  $\mathcal{X} = \{(c,q) \mid c \in x, q \cong c\}$ (e.g., Line~\ref{pre:x}).

During the preprocessing, the set of all negation atoms in $P$, $NANT(P) = \{ a \mid a \in  r^- \land  r \in P\} $~\cite{PontelliSE09,lyly2021}, is computed.
For $P_1$ and the answer set $A = \{n(1), n(2), c, m(1)\}$, we have $NANT(P_1) = \{a,b,c\}$.

\begin{algorithm}[t]
    \DontPrintSemicolon
     
    \KwInput{$D$ - associative array of rules (this is $D_P$), $A$ - an answer set}   
    
    $V_c = \{B \mid D[(0,h)] = B \land h = \emptyset\}$
    
    $E_c \gets \{\emptyset \mapsto \emptyset\}$    \tcp*{Initialize an empty associative array $E_c$: $E_c.key()=\emptyset$}
    
    \For{$B \in V_c$}
    {
        
        $violation \gets \{a \mid  a \in r^+ \land a \in A\} \cup \{\sim\!a \mid a \in r^- \land a \notin A\}$
        
        $support \gets \{\sim\!a \mid  a \in r^+ \land a \notin A\} \cup \{a \mid  a \in r^- \land a \in A\} $ 
        
        \If{choice atom $x$ in $B$ and $ S = \{(c,q) \mid c \in x ,q \cong c, A \models c \land q  \}$}
        {
            $\mathcal{X} = \{(c,q) \mid c \in x, q \cong c\}$  \label{pre:x}

            \If{$x = l~ \{p_1 :q_1, \ldots, p_n:q_n\} ~u \in r^+$ and $|S| < l$ or $|S| > u$} 
            {
                $choice\_support \gets \{``\sim\!(l <= \mathcal{X} <= u)"\}$ 
            }
            \If{$x = l~ \{p_1 :q_1, \ldots, p_n:q_n\} ~u \in r^-$ and $l \leq |S| \leq u$}
            {
                $choice\_support \gets \{``l <= \mathcal{X} <= u"\}$  
            }
            \If{$x = l~ \{p_1 :q_1, \ldots, p_n:q_n\} \in r^+$ or $x = \{p_1 :q_1, \ldots, p_n:q_n\}~ l-1 \in r^-$ and  $|S| < l$}
            {
                $choice\_support \gets \{``\sim\!(l <= \mathcal{X})"\}$ 
            }
            \If{$x = l~ \{p_1 :q_1, \ldots, p_n:q_n\} \in r^-$ or $x = \{p_1 :q_1, \ldots, p_n:q_n\}~ l-1 \in r^+$ and $|S| \geq l$}
            {
                $choice\_support \gets \{``l <= \mathcal{X}"\}$ 
            }
        }

        $support \gets support \cup  choice\_support$
        
        \If{$S \ne \emptyset$}
        {
            $E_c[choice\_support] \gets [S]$
        
            $E_c[p_i] \gets [\{``*True"\}] ~such~that~ p_i \in S$
        }
        \Else
        {
            $E_c[choice\_support] \gets [\{``*Empty"\}]$
        }
        
        \For{$v \in violation$ }
        {   
            
            Append $\{triggered\_constraint(v)\}$ to a list $E_c[v]$  \label{pre:triggeredassign}
            
            $E_c[triggered\_constraint(v)] \gets [ c \cup \{s\} \mid s \in support, c \in E_c[triggered\_constraint(v)]]$ \label{pre:triggeredsupport}
            
        }
    }
    \KwRet $E_c$
    \;
\caption{$constraint\_preprocessing(D,A)$}    
\label{alg:constraintpreprocessing}
\end{algorithm}

\begin{example} \label{example:e}
Let us reconsider the program $P_1$  and its answer set $A = \{n(1), n(2), c, m(1)\}$.  The output of the preprocessing, $E_{r(P_1)}$ (left) and $E_{c(P_1)}$ (right),  are as follows:

\begin{minipage}{0.37\textwidth}
\begin{lstlisting}[numbers=none,escapechar=\%]
%$E_{r(P_1)}$% = {
c    : [{~a}],
~a   : [{c}],
~b   : [{~a}],
m(1) : 
[{c, +choice, n(1)}],
~m(2): 
[{c, -choice, n(2)}],
n(1) : [{T}],
n(2) : [{T}]
}
\end{lstlisting}
\end{minipage}%
\hfill
\begin{minipage}{0.57\textwidth}
\begin{lstlisting}[numbers=none,escapechar=\%]
%$E_{c(P_1)}$% = {
m(1) : [{triggered_constraint(m(1))}],
triggered_constraint(m(1)) : [{~b}],
c : [{triggered_constraint(c)}],
triggered_constraint(c) : 
[{1<={(m(1), n(1)), (n(2), m(2))}<=1}],
1<={(m(1), n(1)), (n(2), m(2))}<=1 : 
[{(m(1), n(1))}],
(m(1), n(1)) : [{*True}]
}
\end{lstlisting}
\end{minipage}%

$E_{r(P_1)}$ shows that
the supported set of two choice heads $m(1)$ and $m(2)$ contains ${+}choice$ and ${-}choice$, respectively, which depends on their truth values and the value of their bodies.

$E_{c(P_1)}$ shows that
atom $b$ in $r_4$ makes the constraint satisfied while $m(1)$ does not support the constraint. Thus, $\{\sim\!b\}$ is the support set of $triggered\_constraint(m(1))$, and $\{triggered\_constraint(m(1))\}$ is the support set of $m(1)$.
For the additional constraint of $P_1$, $l(9)$ is true (encoded in $(m(1),n(1))$) w.r.t. $A$, resulting the constraint is satisfied. 
The truth value of $c$ does not contribute to making the constraint satisfied. 
Thus, $triggered\_constraint(c)$ is added to the explanation of $c$.
\end{example}

\subsection{Minimal assumption set}
\label{subsec:minimalassumption}

\begin{wrapfigure}[11]{r}{0.35\linewidth}
  \centering
  \vspace*{-40pt}
 \includegraphics[width=0.9\linewidth]{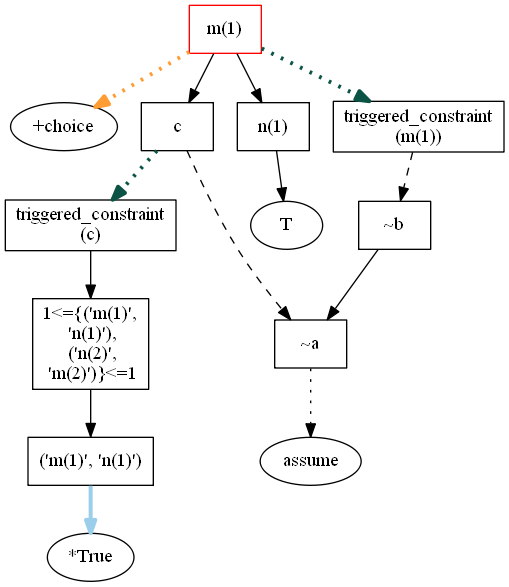}
 \vspace{-10pt}
\caption{\label{fig:m(1)} \scriptsize Explanation of $m(1)$}
\end{wrapfigure}

The idea of the algorithm for computing minimal assumption sets is as follows.
\begin{itemize}[noitemsep,topsep=1pt]
    \item A tentative assumption set $TA$ \cite{PontelliSE09,lyly2021} is computed, which is the superset of minimal assumption sets. 

    \item  $E_{r(P)}$ in Sec.~\ref{subsec:preprocessing} is utilized to compute all derivation paths $M$ of $a \in TA$.
    Then, the derivation paths $M$ are examined whether they satisfy the cycle condition in the definition of explanation graph. During the examination of a derivation path $N \in M$, the other tentative assumption atoms derived from $a$ are stored in a set $D$, which is appended to 
    $DA[a]$ 
    ($DA$ is an associative array)
    if $N$ is satisfied. 
    If $a$ is derivable from other atoms in $TA$, then the relation of $a$ will be checked later in the next step and $a$ is stored in a set $T'$. A set $T=TA \setminus T'$ contains atoms that must assume to be false.
\end{itemize}

\begin{itemize}[noitemsep,topsep=0pt]
    \item We calculate a set of minimal atoms, $min(B)$, that breaks all cycles among tentative assumption atoms via $DA$. Note that the number of set $min(B)$ can be more than one.
    \item Finally, the minimal assumption set $U = T \cup min(B)$. 
\end{itemize}

 \begin{example} \label{example:u}
 Let us reconsider the program $P_1$ and Example \ref{example:e}. 
    \begin{itemize}[noitemsep,topsep=1pt]
     \item For the program $P_1$, we have: 	$TA  = \{a,b\}$
    
     \item From $E_{r(P_1)}$ in Example \ref{example:e}, atom $a$ is not derivable
     from other atoms in $TA$ while atom $b$ is derivable from an atom in $\{a\}$. Thus, we have $T' =\{b\}$, $T = \{a\}$ and $DA = {b : [\{a\}]}$. Also, there is no cycle between $a$ and $b$, so $min(B) = \emptyset$. As a result, the minimal assumption set is $U = \{a\}$.
    \end{itemize}
\end{example}

\subsection{ASP-based explanation system}
\label{subsec:exp}

\begin{wrapfigure}[14]{r}{0.4\linewidth}
  \centering
   \vspace*{-30pt}
 \includegraphics[width=0.9\linewidth]{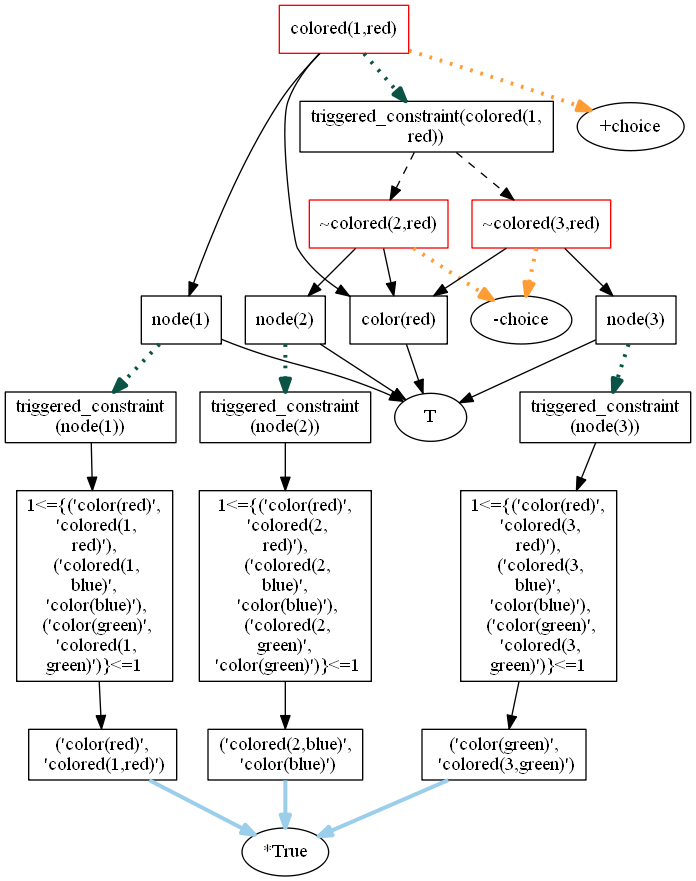}
 \vspace{-10pt}
\caption{\label{fig:choicegraphcoloring} \scriptsize Explanation graph of $colored(1,red)$}
\end{wrapfigure}

In this section, we describe how the explanation graph is generated by utilizing $E_r$, $E_c$ from Sec.~\ref{subsec:preprocessing} and the minimal assumption set $U$ from Sec.~\ref{subsec:minimalassumption}. 
Let $E = \{k \mapsto V \mid k \in E_r.keys() \cup E_c.keys(), V = [r \cup c] ~such~that~ r \in E_r[k] ~and~ c \in E_c[k]
\}$. Note that $r = \emptyset$ if $\nexists k \in E_r.keys()$ and $c = \emptyset$ if $\nexists k \in E_c.keys()$.
Given $E$, the algorithm from \cite{lyly2021} will find the explanation graph of literal in $P$, taking into consideration the additional types of nodes and links. 
\begin{example}
For program $P_1$, the explanation graph of $m(1)$ is shown in Fig. \ref{fig:m(1)}. 

In Fig.~\ref{fig:m(1)}, a justification for $m(1)$ depends positively on $c$ and $n(1)$. A choice head $m(1)$ is chosen to be true. The constraint containing $m(1)$ is satisfied by $A$ because of the truth value of $b$. The constraint containing $c$ is satisfied by $A$ because the conjunction of $n(1)$ and $m(1)$ is true. 
\end{example}

\subsection{Illustration}
\label{sec:illustration}

We illustrate the application of our updated system, \easp{}, to the graph coloring problem. 
We use  a solution of the problem where each node is assigned a unique color by the choice rule:
$1 \{ colored(X,C) : color(C) \} 1 \leftarrow node(X)$.

 Fig.~\ref{fig:choicegraphcoloring} shows the explanation graph of $colored(1,red)$. 
Unlike Fig.~\ref{fig:graphcoloring}, Fig.~\ref{fig:choicegraphcoloring} shows that a choice head $colored(1,red)$ is chosen to be true while two choice heads, $colored(3,red)$ and $colored(2,red)$, are chosen to be false, which are represented via orange dotted  links (link $\bullet$). Fig.~\ref{fig:choicegraphcoloring} displays the constraint that $node(1)$ must assign a different color with $node(3)$ and $node(2)$. This shows via
the links from $colored(1,red)$ to $\sim\!(colored(2,red)$ and $\sim\!(colored(3,red)$ connected through $triggered\_constraint\\(colored(1,red))$ (green dotted  link $\diamond$).
Also, the triggered constraints of each $node(1)$, $node(2)$ and $node(3)$, each node is assigned exactly one color, are shown via the aggregate functions in the node labels (blue solid link $\oplus$).

\section{Conclusion}
In this paper, we proposed an extension of our explanation generation system for ASP programs, {\tt exp(A\\SP$^c$)}, which supports choice rules and includes constraint information.  
At this stage, we have focused on developing an approach that would simplify the program debugging task and have left a systematic investigation of its scalability for a later phase. Nonetheless, in preliminary evaluations, we successfully tested our approach on programs of growing complexity, including one from a  practical application that contains 421 rules, 657 facts and 
has a tentative assumption set of size 44.
An additional future goal is to extend \easp{} so that it can deal with other {\small \tt clingo} constructs like the aggregates $\#sum$, $\#min$,  etc.

\bibliographystyle{eptcs}
\bibliography{generic,bibfile}

\end{document}